\documentclass{article} 
\usepackage{iclr2023_conference,times}
\iclrfinalcopy

\usepackage{amsmath,amsfonts,bm}

\def\eqref#1{equation~\ref{#1}}

\def\1{\bm{1}}

\def\mA{{\bm{A}}}

\def\mC{{\bm{C}}}
\def\mD{{\bm{D}}}

\def\mX{{\bm{X}}}

\DeclareMathAlphabet{\mathsfit}{\encodingdefault}{\sfdefault}{m}{sl}
\SetMathAlphabet{\mathsfit}{bold}{\encodingdefault}{\sfdefault}{bx}{n}

\usepackage[utf8]{inputenc} 
\usepackage[T1]{fontenc}    
\usepackage{times}

\usepackage{hyperref}       
\usepackage{url}            
\usepackage{booktabs}       
\usepackage{amsfonts}       
\usepackage{nicefrac}       
\usepackage{microtype}      
\usepackage{colortbl}
\usepackage{xcolor}
\usepackage{environ}
\usepackage{hyperref}
\usepackage{url}
\usepackage{multicol}
\usepackage{multirow}
\usepackage{amsthm}
\usepackage{graphicx}
\usepackage{tikz}
\usepackage{amsmath}
\usepackage{wrapfig}
\usepackage{makecell}
\usepackage{diagbox}
\usepackage{enumitem}
\usepackage{pifont}
\usepackage{diagbox}
\usepackage{footnote}
\usepackage{algpseudocode}
\usepackage{algorithm2e}
\usepackage{listings}
\usepackage{makecell}
\usepackage{wasysym}
\usepackage{marvosym}
\usepackage[alpine]{ifsym}
\usepackage[normalem]{ulem}
\usepackage{xspace}
\RestyleAlgo{ruled}

\definecolor{codeblue}{RGB}{35, 86, 141}
\definecolor{codepurple}{RGB}{136, 23, 152}
\definecolor{codegray}{RGB}{128, 128, 128}
\definecolor{codeorange}{RGB}{230, 126, 34}
\definecolor{codebrown}{RGB}{88, 41, 0}
\definecolor{codered}{RGB}{220, 50, 47}
\definecolor{azure}{rgb}{0.0, 0.5, 1.0}
\definecolor{bradiesblue}{rgb}{0.0, 0.44, 1.0}
\definecolor{ncsblue}{rgb}{0.0, 0.53, 0.74}
\definecolor{pigment}{rgb}{0.2, 0.2, 0.6}
\definecolor{Blue}{rgb}{0.0, 0.0, 1.0}
\definecolor{Brown}{rgb}{0.65, 0.16, 0.16}
\definecolor{mycite}{cmyk}{0.55,1,0,0.15}

\hypersetup{
    colorlinks=true,
    citecolor=mycite,
    urlcolor=Brown,
    linkcolor=Brown
}

\makeatletter
\def\part{\par
   \addvspace{4ex}%
   \@afterindentfalse
   \secdef\@part\@spart}%
\def\@part[#1]#2{%
 \@ifnum{\c@secnumdepth >\m@ne}{%
        \refstepcounter{part}%
        \addcontentsline{toc}{part}{\thepart\hspace{1em}#1}%
 }{%
      \addcontentsline{toc}{part}{#1}%
 }%
   \nobreak
   \vskip 3ex
   \@afterheading
}%
\makeatother

\newcolumntype{g}{>{\columncolor{codegray!25}}r}
\def\ourmethod{{SimTeG\xspace}}
\def\ourcode{{\small\url{https://github.com/vermouthdky/SimTeG}}}
\def\ourfeature{{\small\url{https://huggingface.co/datasets/vermouthdky/SimTeG}}}

\newcounter{observationcount}
\newcommand{\observation}{%
    \stepcounter{observationcount}%
    \textbf{Observation \arabic{observationcount}: }%
}

\title{SimTeG: A Frustratingly \underline{Sim}ple Approach Improves \underline{Te}xtual \underline{G}raph Learning}

\makeatletter
\renewcommand\@fnsymbol[1]{\ifcase#1\or§\or¶\or††\or‡‡\or§§\or¶¶\or***\else\@ctrerr\fi}
\makeatother

\author{
   Keyu Duan\textsuperscript{1,2*}\thanks{\textsuperscript{*}Correspondence to Keyu Duan (\url{k.duan@u.nus.edu})}, Qian Liu\textsuperscript{2}, Tat-Seng Chua\textsuperscript{1},
   Shuicheng Yan\textsuperscript{2}, Wei Tsang Ooi\textsuperscript{1\dag}\thanks{\textsuperscript{\dag}Equal advising}, \\ 
   \textbf{\ Qizhe Xie\textsuperscript{1\dag}, Junxian He\textsuperscript{3\dag}} \\
   \textsuperscript{1} National University of Singapore
   \textsuperscript{2} Sea AI Lab, Singapore \\
   \textsuperscript{3} The Hong Kong University of Science and Technology \\
}

\begin{document}

\maketitle

\begin{abstract}
   Textual graphs (TGs) are graphs whose nodes correspond to text (sentences or documents), which are widely prevalent.
   The representation learning of TGs involves two stages: \((i)\) \textit{unsupervised feature extraction} and
   \((ii)\) \textit{supervised graph representation learning}.
   In recent years, extensive efforts have been devoted to the latter stage, where Graph Neural Networks (GNNs) have dominated.
   However, the former stage for most existing graph benchmarks still relies on traditional feature engineering techniques.
   More recently, with the rapid development of language models (LMs),
   researchers have focused on leveraging LMs to facilitate the learning of TGs,
   either by jointly training them in a computationally intensive framework (\textit{merging the two stages}), or
   designing complex self-supervised training tasks for feature
   extraction (\textit{enhancing the first stage}). In this work, we present
   \ourmethod, a frustratingly \underline{Sim}ple approach for \underline{Te}xtual \underline{G}raph learning
   that does not innovate in frameworks, models, and tasks.
   Instead, we first perform \emph{supervised} parameter-efficient fine-tuning (PEFT) on a pre-trained
   LM on the downstream task, such as node classification. We then
   generate node embeddings using the last hidden states of finetuned LM. These derived features
   can be further utilized by any GNN for training on
   the same task. We evaluate our approach on two fundamental graph
   representation learning tasks: \textit{node classification} and
   \textit{link prediction}. Through extensive experiments, we show that our approach
   significantly improves the performance of various GNNs on multiple graph
   benchmarks. Remarkably, when additional supporting text provided by large language models
   (LLMs) is included,
   a simple two-layer GraphSAGE trained on an ensemble of \ourmethod\ achieves an accuracy of 77.48\% on \texttt{OGBN-Arxiv}, 
   comparable to state-of-the-art (SOTA) performance obtained from far more complicated GNN architectures. Furthermore, when combined with a SOTA GNN,
   we achieve a new SOTA of \(78.04 \%\) on \texttt{OGBN-Arxiv}.
   Our code is publicly available at \ourcode\ and the
   generated node features for all graph benchmarks can be accessed at \ourfeature.
\end{abstract}

\section{Introduction}\label{sec:intro}
Textual Graphs (TGs) offer a graph-based representation of
text data where relationships between phrases, sentences, or documents are depicted through edges.
TGs are ubiquitous in real-world applications, including citation graphs~\citep{hu2020open,yang2016revisiting},
knowledge graphs~\citep{wang2021wikigraphs}, and social networks~\citep{zeng2019graphsaint,hamilton2017inductive},
provided that each entity can be represented as text. Different from traditional NLP tasks, instances in TGs
are correlated with each other, which provides non-trivial and specific information for downstream tasks.
In general, graph benchmarks are usually task-specific~\citep{hu2020open}, and most TGs are designed for two fundamental
tasks: \textit{node classification} and \textit{link prediction}. For the first one, we aim to predict the category of
unlabeled nodes while for the second one, our goal is to predict missing links among nodes. For both tasks, text attributes
offer critical information.


In recent years, TG representation learning follows a two-stage paradigm: \((i)\) \textit{upstream: unsupervised feature extraction} that encodes
text into numeric embeddings, and \((ii)\) \textit{downstream: supervised graph representation learning} that further transform the embeddings utilizing the graph structure.
While Graph Neural Networks (GNNs) have dominated the latter stage, with an extensive body of academic research published,
the former stage surprisingly still relies on traditional feature engineering techniques. For example,
in most existing graph benchmarks~\citep{hu2020open,yang2016revisiting,zeng2019graphsaint}, node features are constructed
using bag-of-words (BoW) or skip-gram~\citep{mikolov2013distributed}. This intuitively limits the performance of downstream GNNs, as
it fails to fully capture textual semantics, fostering an increasing number of GNN models with more and more complex structures.
More recently, researchers have begun to leverage the power of language models (LMs) for TG representation learning.
Their efforts involve \((i)\) designing complex tasks for LMs to
generate powerful node representations~\citep{chien2021node}; \((ii)\) jointly training LMs and GNNs in a specific framework~\citep{zhao2022learning,mavromatis2023train};
or \((iii)\) fusing the architecture of LM and GNN for end-to-end training~\citep{yang2021graphformers}. These works focus on novel
\textit{training tasks}, \textit{model architectures}, or \textit{training frameworks}, which generally require substantially modifying the training procedure. However, we argue that such complexity is not actually needed. As a response, in this paper, we present a frustratingly simple yet
highly effective method that does not innovate in any of the above aspects but significantly improves the performance of GNNs on TGs.

We are curious about several research questions: \((i)\) How much could language models' features \textit{generally} improve the
learning of GNN: is the improvement specific for certain GNNs? \((ii)\) What kind of language models fits the needs of textual graph
representation learning best? \((iii)\) How important are text attributes on various graph tasks: though
previous efforts have shown improvement in node classification, is it also beneficial for link prediction, an equivalently fundamental task that
intuitively emphasizes more on graph structures?
To the end of answering the above questions, we take an initial step forwards by introducing a simple,
effective, yet surprisingly neglected method on TGs and empirically evaluating it on two fundamental graph tasks:
node classification and link prediction. Intuitively, when
omitting the graph structures, the two tasks are equivalent to text classification and text similarity (retrieval) tasks in NLP,
respectively. This intuition motivates us to propose our method: \ourmethod.
We first parameter-efficiently finetune (PEFT) an LM on the textual corpus of a TG with task-specific labels and then use the finetuned LM to
generate node representations given its text by removing the head layer. Afterward, a GNN is trained with the derived node embeddings
on the \textit{same} downstream task for final evaluation. Though embarrassingly simple, \ourmethod\ shows remarkable performance on
multiple graph benchmarks w.r.t. node classification and link prediction. Particularly, we find several \textit{key observations}:

\ding{182} Good language modeling could generally improve the learning of GNNs on both node classification and link prediction. We evaluate
\ourmethod\ on three prestigious graph benchmarks for either node classification or link prediction, and find that \ourmethod\
consistently outperforms the official features and the features generated by pretrained LMs (without finetuning) by a large margin.
Notably, backed with SOTA GNN, we achieve \textit{new SOTA performance} of \(78.02 \%\) on \texttt{OGBN-Arxiv}.
See Sec.~\ref{sec:main_exp} and Appendix~\ref{sec:more_exp_results} for details.

\ding{183} \ourmethod\ significantly complements the margin between GNN backbones on multiple graph benchmarks
by improving the performance of \emph{simple} GNNs.
Notably, a simple two-layer GraphSAGE~\citep{hamilton2017inductive} trained on \ourmethod\ with proper LM backbones achieves
on-par SOTA performance of \(77.48 \%\) on \texttt{OGBN-Arxiv}~\citep{hu2020open}.

\ding{184} PEFT are crucial when finetuning LMs to generate representative embeddings, because full-finetuning usually leads to extreme overfitting
due to its large parameter space and the caused fitting ability. The overfitting in the LM finetuning stage will hinder the training of
downstream GNNs with a collapsed feature space. See Sec.~\ref{sec:ablation_peft} for details.

\ding{185} \ourmethod\ is moderately sensitive to the selection of LMs. Generally, the performance of \ourmethod\ is positively correlated with
the corresponding LM's performance on text embedding tasks, e.g. classification and retrieval.
We refer to Sec.~\ref{sec:ablation_lm} for details. Based on this, we expect further improvement of \ourmethod\ once more powerful LMs for
text embedding are available.

\vspace{-3mm}
\section{Related Works}\label{sec:related_works}
\vspace{-3mm}
In this section, we first present several works that are closely related to ours and further clarify
several concepts and research lines that are plausibly related to ours in terms of similar terminology.

\noindent
\textbf{Leveraging LMs on TGs.} Focusing on leveraging the power of LMs to TGs, there are several works
that are existed and directly comparable with ours. For these works, they either focus on
\((i)\) \textit{designing specific strategies to generate node embeddings using LMs}~\citep{he2023explanations,chien2021node} or \((ii)\)
\textit{jointly training LMs and GNNs within a framework}~\citep{zhao2022learning,mavromatis2023train}.
Representatively, for the former one, \citet{chien2021node} proposed a self-supervised graph learning task integrating XR-Transformers~\citep{zhang2021fast}
to extract node representation, which shows superior performance on multiple graph benchmarks, validating the necessity for acquiring
high-quality node features for attributed graphs.
Besides, \citet{he2023explanations} utilizes ChatGPT~\citep{john2023chatgpt} to generate additional supporting text
with LLMs. For the latter mechanism, \citet{zhao2022learning} proposed a variational expectation maximization
joint-training framework for LMs and GNNs to learn powerful graph representations. \citet{mavromatis2023train}
designs a graph structure-aware framework to distill the knowledge from GNNs to LMs. Generally, the joint-training framework
requires specific communication between LMs and GNNs, e.g. pseudo labels~\citep{zhao2022learning} or hidden states~\citep{mavromatis2023train}.
It is worth noting that the concurrent work \citet{he2023explanations} proposed a close method to ours.
However, \citet{he2023explanations} focuses on generating additional informative texts for nodes with LLMs,
which is specifically for citation networks on node classification task. In contrast, we focus on generally investigating
the effectiveness of our proposed method, which could be widely applied to unlimited datasets and tasks. Utilizing the additional
text provided by \citet{he2023explanations}, we further show that our method could achieve now SOTA on \texttt{OGBN-Arxiv}.
In addition to the main streams, there are related works trying to fuse the architecture of LM and GNN for end-to-end
training. \citet{yang2021graphformers} proposed a nested architecture by injecting GNN layers into LM layers. However, due to the natural
incompatibleness regarding training batch size, this architecture only allows \(1\)-hop message passing, which significantly
reduce the learning capability of GNNs.


\noindent
\textbf{More ``Related'' Works.} \ding{182} \textit{Graph Transformers}~\citep{wu2021representing,
   ying2021transformers,hussain2022global,park2022grpe,chen2022structure}: Nowadays, Graph Transformers
are mostly used to denote Transformer-based architectures that embed both topological structure and node features.
Different from our work, these models focus on graph-level problems (e.g. graph classification and graph
generation) and specific domains (e.g. molecular datasets and protein association networks), which cannot
be adopted on TGs. \ding{183} \textit{Leveraging GNNs on Texts}~\citep{zhu2021textgnn,huang2019text,zhang2020every}:
Another seemingly related line on integrating GNNs and LMs is conversely applying GNNs to textual documents.
Different from TGs, GNNs here do not rely on ground-truth graph structures but the self-constructed or synthetic ones.

\vspace{-2mm}
\section{Preliminaries}\label{sec:preliminaries}
\vspace{-2mm}
\noindent
\textbf{Notations.} To make notations consistent, we use \textbf{bold} uppercase letters to denote matrices and vectors,
and calligraphic font types (e.g. \(\mathcal{T}\)) to denote sets.
We denote a textual graph as a set \(\mathcal{G}=(\mathcal{V}, \mathcal{E}, \mathcal{T})\),
where \(\mathcal{V}\) and \(\mathcal{E}\) are a set of nodes and edges, respectively.
\(\mathcal{T}\) is a set of text and each textual item is aligned with a node
\(v \in \mathcal{V}\). For practical usage, we usually rewrite \(\mathcal{E}\)
into \(\mathbf{A} \in \{0, 1\}^{|\mathcal{V}| \times |\mathcal{V}|}\), which is a sparse matrix,
where entry \(\mathbf{A}_{i,j}\) denotes the link between node
\(v_i, v_j \in \mathcal{V}\).

\noindent
\textbf{Problem Formulations.}
We focus on two fundamental tasks in TGs: \((i)\) \textit{node classification}
and \((ii)\) \textit{link prediction}. For node classification, given a TG \(\mathcal{G}\),
we aim to learn a model \(\Phi : \mathcal{V} \rightarrow \mathcal{Y}\),
where \(\mathcal{Y}\) is the ground truth labels. For link prediction, given a TG \(\mathcal{G}\),
we aim to learn a model \(\Phi: \mathcal{V} \times \mathcal{V} \rightarrow \{0, 1\}\),
where \(f(v_i, v_j) = 1\) if there is a link between \(v_i\) and \(v_j\), otherwise \(f(v_i, v_j) = 0\).
Different from traditional tasks that are widely explored by the graph learning community,
evolving original text into learning is non-trivial. Particularly, when ablating the graphs structure,
node classification and link prediction problem are collapsed to text classification and
text similarity problem, respectively. This sheds light on how to leverage LMs for TG representation
learning.

\noindent
\textbf{Node-level Graph Neural Networks.}\label{sec:prelim_gnn} 
Nowadays, GNNs have dominated graph-related tasks. Here we focus on GNN models working on node-level
tasks (i.e. \textit{node classification} and \textit{link prediction}).
These models work on generating node representations by recursively aggregating features from their
multi-hop neighbors, which is usually noted as \textit{message passing}. Generally, one can formulate a
graph convolution layer as: \(\mX_{l+1} = \Psi_l(\mC\mX_{l})\), where \(\mC\) is the graph convolution matrix
(e.g. \(\mC = \mD^{-1/2}\mA\mD^{-1/2}\) in Vanilla GCN~\citep{kipf2016semi}) and \(\Psi_l\) is the feature
transformation matrix. For the node classification problem, a classifier (e.g., an MLP) is usually appended to
the output of a \(k\)-layer GNN model; while for link prediction, a similarity function is applied to
the final output to compute the similarity between two node embeddings. As shown above, as GNNs
inherently evolve the whole graph structure for convolution, it is notoriously challenging for scaling it up.
It is worth noting that evolving sufficient
neighbors during training is crucial for GNNs. Many studies~\citep{duan2022comprehensive,zou2019layer}
have shown that full-batch training generally outperforms mini-batch for GNNs on multi graph benchmarks.
In practice, the lower borderline of batch size for training GNNs is usually thousands. However,
when applying it to LMs, it makes the GNN-LM end-to-end training intractable, as a text
occupies far more GPU memories than an embedding.

\noindent
\textbf{Text Embeddings and Language Models.}\label{sec:prelim_lm} 
Transforming text in low-dimensional dense embeddings serves as the upstream of textual graph representation learning and
has been widely explored in the literature.
To generate sentence embeddings with LMs, two commonly-used methods are \((i)\) average
pooling~\citep{reimers2019sentence} by taking the average of all word embeddings
along with attention mask and \((ii)\) taking the embedding of the \texttt{[CLS]}
token~\citep{devlin2018bert}.
With the development of pre-trained language
models~\citep{devlin2018bert,liu2019roberta}, particular language models~\citep{li2020sentence,reimers2019sentence}
for sentence embeddings have been proposed and shown promising results in various
benchmarks~\citep{muennighoff2022mteb}.

\section{\ourmethod: Methodology}\label{sec:methodology}

\begin{figure}[!t]
   \vspace{-3mm}
   \centering
   \includegraphics[width=\textwidth]{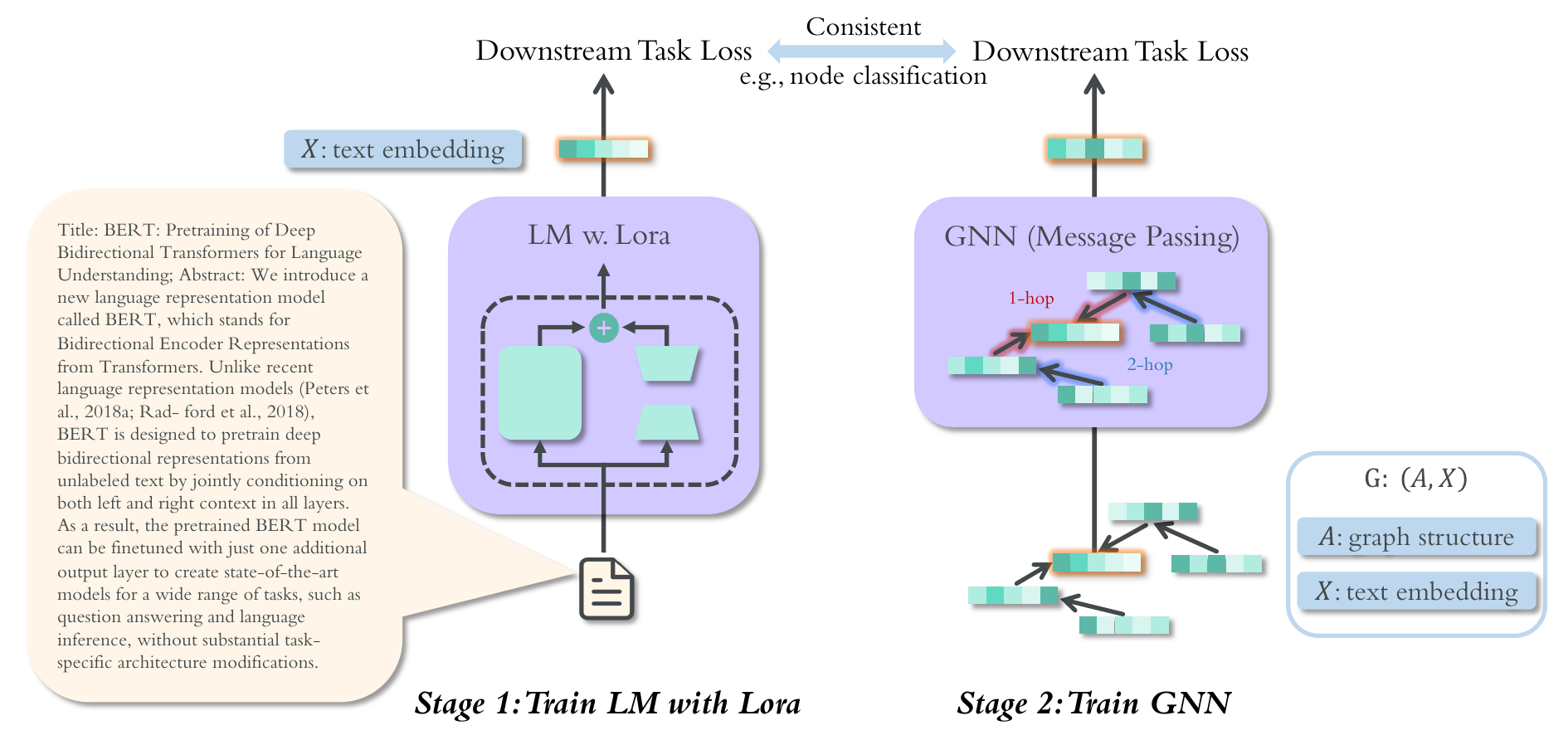}
   \caption{The overview of \ourmethod. In \textit{stage 1}, we train a LM with lora~\citep{hu2022lora} and then
      generate the embeddings \(\mX\) as the representation of text. In \textit{stage 2},
      we train a GNN on top of the embeddings \(\mX\), along with the graph structure. The two stages are guided
      with consistent loss function, e.g., link prediction or node classification.}\label{fig:architecture}
   \vspace{-5mm}
\end{figure}

We propose an extremely simple two-stage training manner that
decouples the training of \(gnn(\cdot)\) and \(lm(\cdot)\). We first finetune \(lm\) on \(\mathcal{T}\)
with the downstream task loss:
\begin{equation}
   Loss_{cls} = \mathcal{L}_{\theta}\big(\phi(lm(\mathcal{T})), \mathbf{Y}\big),
   \quad Loss_{link} = \mathcal{L}_{\theta}\big(\phi\big(lm(\mathcal{T}_{src}), lm(\mathcal{T}_{dst})\big), \mathbf{Y} \big),
\end{equation}
where \(\phi(\cdot)\) is the classifier (left for \textit{node classification}) or
similarity function (right for \textit{link prediction}) and \(\mathbf{Y}\) is the label.
After finetuning, we generate node representations \(\mX\)
with the finetuned LM \(\hat{lm}\). In practice, we follow \citet{reimers2019sentence} to perform mean pooling
over the output of the last layer of the LM and empirically find that such a strategy is more stable and converges faster than
solely taking the \texttt{<cls>} token embedding as representation~\citep{zhao2022learning}.
In the second stage, we train \(gnn\) on \((\mA, \mX)\) with the same task. The corresponding loss is computed
by replacing \(lm(\mathcal{T})\) with \(gnn(\mA, \mX)\).
The two stage is fully decoupled and one can take advantage of any existing GNN and LM models.
We illustrate the two stages in Fig.~\ref{fig:architecture} and the pseudo code is presented in Appendix~\ref{sec:pseudo_code}.

\noindent
\textbf{Regularization with PEFT.} When fully finetuning a LM, the inferred features are prone
to overfit the training labels, which results in collapsed feature space and thus hindering the generalization
in GNN training. Though PEFT was proposed to accelerate the finetuning process without loss of performance, in our
two-stage finetuning stage, we empirically find PEFT~\citep{hu2022lora,houlsby2019parameter,he2022towards}
could alleviate the overfitting issue to a large extent and thus provide well-regularized node features.
See Sec.~\ref{sec:ablation_peft} for empirical analysis. In this work,
We take the popular PEFT method, lora~\citep{hu2022lora}, as the instantiation.

\textbf{Selection of LM.} As the revolution induced by LMs, a substantial number of valuable pre-trained LMs have been proposed. As mentioned before,
when ablating graph structures of TG, the two fundamental tasks, \textit{node classification} and \textit{link prediction}, are
simplified into two well-established NLP tasks, \textit{text classification} and \textit{text similarity (retrieval)}. Based on this
motivation, we select LMs pretrained for information retrieval as the backbone of \ourmethod.
Concrete models are selected based on the benchmark MTEB\footnote{\url{https://huggingface.co/spaces/mteb/leaderboard}}
considering the model size and the performance on both retrieval and classification tasks.
An ablation study regarding this motivation is presented in Sec.~\ref{sec:ablation_lm}.

\begin{figure}[!t]
   \centering
   \includegraphics[width=\textwidth]{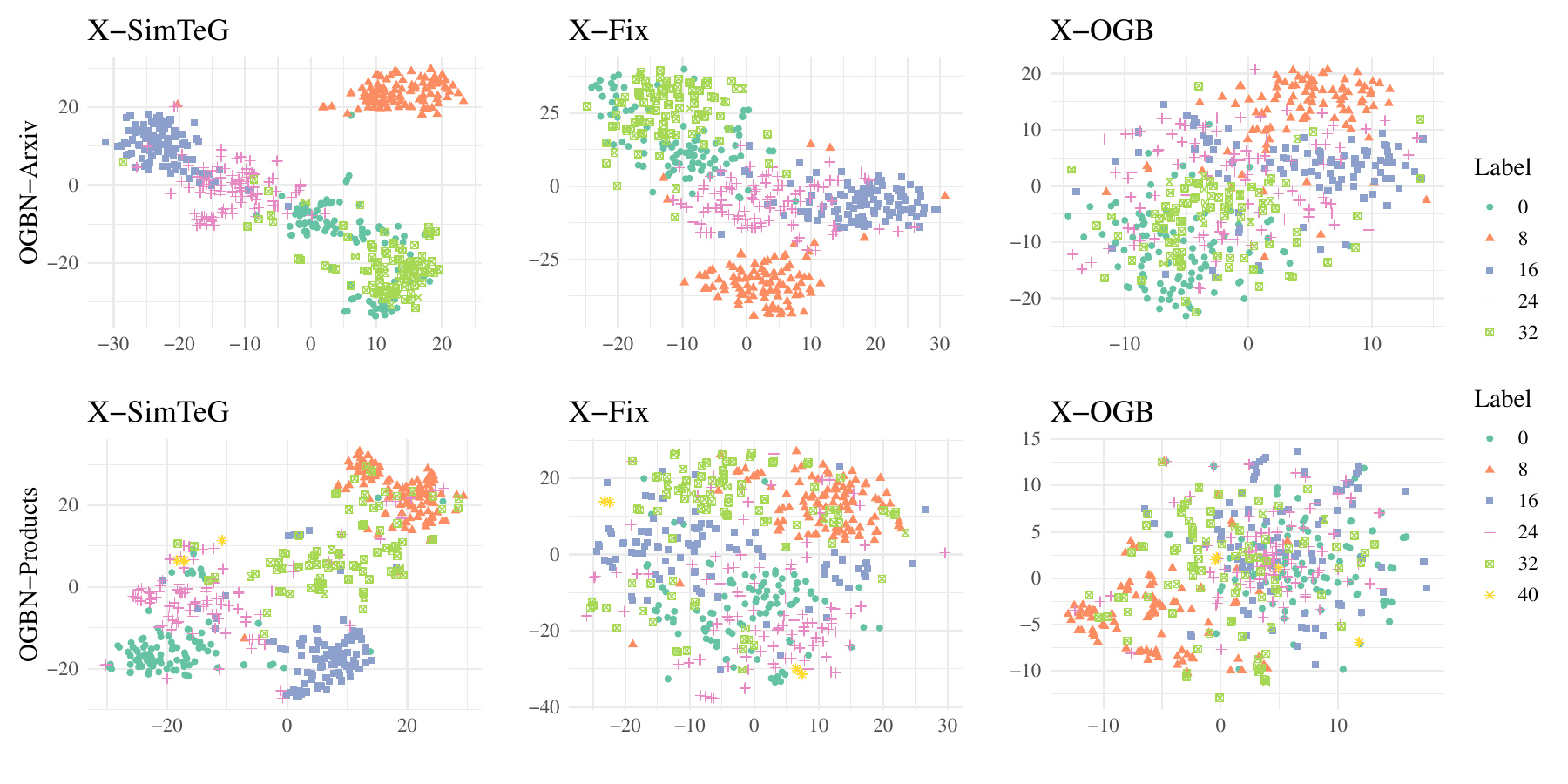}
   \vspace{-6mm}
   \caption{The two-dimensional feature space of \(\mX\)-\ourmethod, \(\mX\)-Fix, and \(\mX\)-OGB
      for \texttt{OGBN-Arixv}, and \texttt{OGBN-Products}. different values and shapes refer
      to different labels on the specific dataset. The feature values are computed by T-SNE.
      The LM backbone is e5-large.}\label{fig:feature_space}
\end{figure}

\noindent
\textbf{A Finetuned LM Provides A More Distinguishable Feature Space.}
We plot the two-dimensional feature space computed by T-SNE~\citep{van2008visualizing} of \(\mX\)-\ourmethod, \(\mX\)-Fix (features generated by pretrained LM without finetuning),
and \(\mX\)-OGB regarding labels on
\texttt{OGBN-Arxiv} and \texttt{OGBN-Products} in Fig.~\ref{fig:feature_space}.
In detail, we randomly select 100 nodes each with various labels and use T-SNE to compute its two-dimensional features.
As shown below, \(\mX\)-\ourmethod\ has a significantly more distinguishable feature space as it captures more semantic information and is finetuned
on the downstream dataset. Besides, we find that \(\mX\)-Fix is more distinguishable than \(\mX\)-OGB, which illustrates the
inner semantic capture ability of LMs. Furthermore, in comparison with \texttt{OGBN-Arixv}, features in \texttt{OGBN-Products}
is visually indifferentiable, indicating the weaker correlation between semantic information and task-specific labels. It
accounts for the less improvement of \ourmethod\ on \texttt{OGBN-Products} in Sec.~\ref{sec:main_exp}.

\section{Experiments}\label{sec:empirical_analysis}

In the experiments, we aim at answering three research questions as proposed in the introduction (Sec.~\ref{sec:intro}).
For a clear statement, we split and reformat them into the following research questions.
\noindent
\textbf{Q1:} How much could \ourmethod\ generally improve the learning of GNNs on node classification and link prediction?
\textbf{Q2:} Does X-\ourmethod\ facilitate better convergence for GNNs?
\textbf{Q3:} Is PEFT a necessity for LM finetuning stage?
\textbf{Q4:} How sensitive is GNN training sensitive to the selection of LMs?

\noindent
\textbf{Datasets.} Focusing on two fundamental tasks node classification and link prediction,
we conduct experiments on three prestigious benchmarks: \texttt{OGBN-Arxiv} (Arxiv), \texttt{OGBN-Products} (Products), and
\texttt{OGBL-Citation2}~\citep{hu2020open}. The former two are for node classification while the latter one
is for link prediction. For the former two, we follow the public split, and all text resources are provided by the officials.
For the latter one, \texttt{OGBL-Citation2}, as no official text resources are provided, we take the intersection of it and another
dataset \texttt{ogbn-papers100M} w.r.t. unified paper ids, which results in a subset of \texttt{OGBL-Citation2} with about 2.7M nodes.
The public split is further updated according to this subset. In comparison, the original \texttt{OGBL-Citation2} has about 2.9M nodes,
which is on par with the TG version, as the public valid and test split occupies solely 2\% overall.
As a result, we expect roughly consistent performance for methods on the TG version of \texttt{OGBL-Citation2} and the original one.
We introduce the
statistics of the three datasets in Table.~\ref{tab:dataset_stats} and the details in
Appendix~\ref{sec:dataset_details}.

\noindent
\textbf{Baselines.} We compare \ourmethod\ with the official features \(\mX\)-OGB~\citep{hu2020open}, which is the mean of word embeddings generated by
skip-gram~\citep{mikolov2013distributed}. In addition, for node classification, we include another two SOTA methods:
\(\mX\)-GIANT~\citep{chien2021node} and GLEM~\citep{zhao2022learning}. Particularly, \(\mX\)-* are methods are different at
learning node embeddings and any GNN model could be applied in the downstream task for a fair comparison.
To make things consistent, we denote our method as \(\mX\)-\ourmethod\ without further specification.

\noindent
\textbf{GNN Backbones.} Aiming at investigating the general improvement of \ourmethod, for each dataset,
we select two commonly-used baselines GraphSAGE and MLP besides one corresponding SOTA GNN models based on the official
leaderboard\footnote{\url{https://ogb.stanford.edu/docs/leader_nodeprop}}. For \texttt{OGBN-Arxiv}, we select RevGAT~\citep{li2021training};
for \texttt{OGBN-Products}, we select SAGN+SCR~\citep{sun2021scalable,zhang2021improving}; and for \texttt{ogbn-citation2}, we select
SEAL~\citep{zhang2018link}.

\noindent
\textbf{LM Backbones.} For retrieval LM backbones, we select three popular LMs on MTEB~\citep{muennighoff2022mteb}
leaderboard\footnote{\url{https://huggingface.co/spaces/mteb/leaderboard}} w.r.t. model size and performance on classification
and retrieval: \textit{all-MiniLM-L6-v2}~\citep{reimers2019sentence}, \textit{all-roberta-large-v1}~\citep{reimers2019sentence}, and
\textit{e5-large-v1}~\citep{wang2022text}. We present the properties of the three LMs in Table.~\ref{tab:lm_stats}.

\noindent
\textbf{Hyperparameter search.}
We utilize \textit{optuna}~\citep{akiba2019optuna} to perform hyperparameter search on all tasks. The search space for LMs and GNNs on
all datasets is presented in Appendix~\ref{sec:hp_search_space}.

\subsection{\textbf{Q1}: How much could \ourmethod\ \textit{generally} improve the
   learning of GNNs on node classification and link prediction?}\label{sec:main_exp}

In this section, we conduct experiments to show the superiority of \ourmethod\ on improving the learning of GNNs on node classification and link prediction.
The reported results are selected based on the validation dataset. We present the results based on \textit{e5-large} backbone in Table.~\ref{tab:main_results}
and present the comprehensive results of node classification and link prediction with all the three selected backbones in Table~\ref{tab:node_cls_results_main}
and Table~\ref{tab:link_pred_results_main}. Specifically, in Table~\ref{tab:main_results}, we present two comparison metric \(\Delta_\textit{MLP}\) and
\(\Delta_\textit{GNN}\) to describe the performance margin of (\textit{SOTA GNN, MLP}) (\textit{SOTA GNN, GraphSAGE}), respectively. The smaller the
value is, even negative, the better the performance of simple models is. In addition, we ensemble the GNNs with multiple node embeddings
generated by various LMs and text resources on \texttt{OGBN-Arxiv} and show the results in Table~\ref{tab:ensembling}.
We find several interesting observations as follows.

\begin{table}[!t]
   \vspace{-5mm}
   \centering
   \caption{The performance of SOTA GNN, GraphSAGE and MLP on \texttt{OGBN-Arxiv}, \texttt{OGBN-Products}, \texttt{OGBL-Citation2},
      which are averaged over \(10\) runs (Please note the we solely train LM once to generate the node embeddings). The results of
      GLEM is from the orignal paper. We \textbf{bold}
      the best results w.r.t. the same GNN backbone and \textcolor{red}{red color} the smallest \(\Delta_\textit{MLP}\) and \(\Delta_\textit{GNN}\).}
   \label{tab:main_results}
   \resizebox{1.0\linewidth}{!}{
      \begin{tabular}{lll|c|cg|cg}
         \toprule
         \multirow{2}{*}{Dataset}   & \multirow{2}{*}{Metric}      & \multirow{2}{*}{Method} & SOTA GNN                    & \multicolumn{4}{c}{A \(2\)-layer Simple MLP / GNN}                                                                                   \\
         \cmidrule(lr){4-4} \cmidrule(lr){5-8}
                                    &                              &                         & RevGAT                      & MLP                                                & \(\Delta_\textit{MLP}\) & GraphSAGE                   & \(\Delta_\textit{GNN}\) \\
         \midrule
         \multirow{4}{*}{Arxiv}     & \multirow{4}{*}{Acc. (\%)}   & \(\mX\)-OGB             & 74.01 \(\pm\) 0.29          & 47.73 \(\pm\) 0.29                                 & 25.24                   & 71.80 \(\pm\) 0.20          & 3.40                    \\
                                    &                              & \(\mX\)-GIANT           & 75.93 \(\pm\) 0.22          & 71.08 \(\pm\) 0.22                                 & 4.85                    & 73.70 \(\pm\) 0.09          & 2.23                    \\
                                    &                              & GLEM                    & 76.97 \(\pm\) 0.19          & -                                                  & -                       & 75.50 \(\pm\) 0.24          & 1.47                    \\
         \cmidrule(lr){3-8}
                                    &                              & \(\mX\)-\ourmethod      & \textbf{77.04 \(\pm\) 0.13} & \textbf{74.06 \(\pm\) 0.13}                        & \textcolor{red}{2.98}   & \textbf{76.84 \(\pm\) 0.34} & \textcolor{red}{0.20}   \\
         \bottomrule
         \toprule
         \multirow{2}{*}{Dataset}   & \multirow{2}{*}{Metric}      & \multirow{2}{*}{Method} & SOTA GNN                    & \multicolumn{4}{c}{A \(2\)-layer Simple MLP / GNN}                                                                                   \\
         \cmidrule(lr){4-4} \cmidrule(lr){5-8}
                                    &                              &                         & SAGN+SCR                    & MLP                                                & \(\Delta_\textit{MLP}\) & GraphSAGE                   & \(\Delta_\textit{GNN}\) \\
         \midrule
         \multirow{4}{*}{Products}  & \multirow{4}{*}{Acc. (\%)}   & \(\mX\)-OGB             & 81.82 \(\pm\) 0.44          & 50.86 \(\pm\) 0.26                                 & 30.96                   & 78.81 \(\pm\) 0.23          & 3.01                    \\
                                    &                              & \(\mX\)-GIANT           & 86.12 \(\pm\) 0.34          & \textbf{77.58 \(\pm\) 0.24}                        & \textcolor{red}{8.54}   & 82.84 \(\pm\) 0.29          & 3.28                    \\
                                    &                              & GLEM                    & \textbf{87.36 \(\pm\) 0.07} & -                                                  & -                       & 83.16 \(\pm\) 0.19          & 4.20                    \\
         \cmidrule(lr){3-8}
                                    &                              & \(\mX\)-\ourmethod      & 85.40 \(\pm\) 0.28          & 76.73 \(\pm\) 0.44                                 & 8.67                    & \textbf{84.59 \(\pm\) 0.44} & \textcolor{red}{0.81}   \\
         \bottomrule
         \toprule
         \multirow{2}{*}{Dataset}   & \multirow{2}{*}{Metric}      & \multirow{2}{*}{Method} & SOTA GNN                    & \multicolumn{4}{c}{A \(2\)-layer Simple MLP / GNN}                                                                                   \\
         \cmidrule(lr){4-4} \cmidrule(lr){5-8}
                                    &                              &                         & SEAL                        & MLP                                                & \(\Delta_\textit{MLP}\) & GraphSAGE                   & \(\Delta_\textit{GNN}\) \\
         \midrule
         \multirow{4}{*}{Citation2} & \multirow{2}{*}{MRR (\%)}    & \(\mX\)-OGB             & 86.14 \(\pm\) 0.40          & 25.44 \(\pm\) 0.01                                 & 60.70                   & 77.31 \(\pm\) 0.90          & 8.83                    \\
                                    &                              & \(\mX\)-\ourmethod      & \textbf{86.66 \(\pm\) 1.21} & \textbf{72.90 \(\pm\) 0.14}                        & \textcolor{red}{13.76}  & \textbf{85.13 \(\pm\) 0.73} & \textbf{1.53}           \\
         \cmidrule(lr){2-8}
                                    & \multirow{2}{*}{Hits@3 (\%)} & \(\mX\)-OGB             & 90.92 \(\pm\) 0.32          & 28.22 \(\pm\) 0.02                                 & 62.70                   & 85.56 \(\pm\) 0.69          & 5.36                    \\
                                    &                              & \(\mX\)-\ourmethod      & \textbf{91.42 \(\pm\) 0.19} & \textbf{80.55 \(\pm\) 0.13}                        & \textcolor{red}{10.87}  & \textbf{91.62 \(\pm\) 0.87} & \textcolor{red}{-0.20}  \\
         \bottomrule
      \end{tabular}
   }
   \vspace{-3mm}
\end{table}

\noindent
\observation
\textbf{\ourmethod\ generally improves the performance of GNNs on node classification and link prediction by a large margin.}
As shown in Table~\ref{tab:main_results}, \ourmethod\ consistently outperforms the original features on all datasets and backbones. Besides, in comparison
with \(\mX\)-GIANT, a LM pretraining method that utilizes the graph structures, \ourmethod\ still achieves better performance on \texttt{OGBN-Arxiv} with all
backbones and on \texttt{OGBN-Products} with GraphSAGE, which further indicates the importance of text attributes per se.

\noindent
\observation
\textbf{(\textit{\(\mX\)-\ourmethod\ + GraphSAGE}) consistently outperforms (\textit{\(\mX\)-OGB + SOTA GNN}) on all the three datasets.}
This finding implies that the incorporation of advanced text features can bypass the necessity of complex GNNs, which is why we perceive our method
to be \textit{frustratingly} simple. Furthermore, when replacing GraphSAGE with the corresponding SOTA GNN in \(\mX\)-\ourmethod,
although the performance is improved moderately, this margin of improvement is notably smaller compared to the performance gap on \(\mX\)-OGB.
Particularly, we show that the simple 2-layer GraphSAGE achieves comparable performance with the dataset-specific SOTA GNNs.
Particularly, on \texttt{OGBN-Arxiv}, GraphSAGE achieves \(76.84 \%\), taking the \textbf{\textit{4-th}} place in the corresponding leaderboard
(by 2023-08-01). Besides, on \texttt{OGBL-Citation2}, GraphSAGE even outperforms the SOTA GNN method SEAL on Hits@3.

\noindent
\observation
\textbf{With additional text attributes, \ourmethod\ with Ensembling achieves \textit{new SOTA performance} on \texttt{OGBN-Arxiv}.}
We further demonstrate the effectiveness of \ourmethod\ by ensembling the node embeddings generated by different LMs and texts.
For text, we use both the original text provided by \citet{hu2020open} and the additional text attributes\footnote{It is worth noting
   that as GPT-4 used by \citet{he2023explanations} does not
   release their training recipe, we do not know whether the arxiv papers are included during training,
   which may lead to a label leakage problem.} provided by \citet{he2023explanations},
which is generated by ChatGPT. For LMs, we use both e5-large and all-roberta-large-v1. We train GraphSAGE or RevGAT on those node embeddings
generated by various LMs and texts, and make the final predictions with weighted ensembling (taking the weighted average of all predictions).
As shown in Table~\ref{tab:ensembling}, with RevGAT, we achieve new SOTA performance on \texttt{OGBN-Arxiv} with 78.03\% test accuracy, more than
0.5 \% higher than the previous SOTA performance (77.50\%) achieved by \citet{he2023explanations}. It further validates the importance of
text features and the effectiveness of \ourmethod.

\begin{table}[!t]
   \centering
   \caption{The performance of GraphSAGE and RevGAT trained on \texttt{OGBN-Arxiv} with additional text attributes
      provided by \citet{he2023explanations}.
      LMs for ensembling are e5-large and all-roberta-large-v1.
      We select the top-3 SOTA methods from the leaderboard of \texttt{OGBN-Arxiv} (accessed on \textit{2023-07-18}) for comparison
      and \colorbox{codegray!25}{gray color} our results (reported over 10 runs).}\label{tab:ensembling}.
   \resizebox{0.95\linewidth}{!}{
      \begin{tabular}{lllcc}
         \toprule
         Rank                       & Method                                      & GNN Backbone                     & Valid Acc. (\%)                                      & Test Acc. (\%)                                     \\
         \midrule
         \cellcolor{codegray!25}1   & \cellcolor{codegray!25}TAPE + SimTeG (ours) & \cellcolor{codegray!25}RevGAT    & \cellcolor{codegray!25}\textbf{{78.46 \(\pm\) 0.04}} & \cellcolor{codegray!25}\textbf{78.03 \(\pm\) 0.07} \\
         2                          & TAPE~\citep{he2023explanations}             & RevGAT                           & 77.85 \(\pm\) 0.16                                   & 77.50 \(\pm\) 0.12                                 \\
         \cellcolor{codegray!25}{3} & \cellcolor{codegray!25}TAPE + SimTeG (Ours) & \cellcolor{codegray!25}GraphSAGE & \cellcolor{codegray!25}77.89 \(\pm\) 0.08            & \cellcolor{codegray!25}77.48 \(\pm\) 0.11          \\
         4                          & GraDBERT~\citep{mavromatis2023train}        & RevGAT                           & 77.57 \(\pm\) 0.09                                   & 77.21 \(\pm\) 0.31                                 \\
         5                          & GLEM~\citep{zhao2022learning}               & RevGAT                           & 77.46 \(\pm\) 0.18                                   & 76.94 \(\pm\) 0.25                                 \\
         \bottomrule
      \end{tabular}
   }
   \vspace{-3mm}
\end{table}

\noindent
\observation
\textbf{Text attributes are unequally important for different datasets.}
As shown in Table~\ref{tab:main_results}, we compute \(\Delta_\textit{MLP}\) which is the performance gap between MLP and SOTA GNNs.
Empirically, this value indicates the importance of text attributes on the corresponding dataset, as MLP is solely trained
on the texts (integrated with SOTA LMs) while SOTA GNN additionally takes advantage of graph structures. Therefore, approximately, the less \(\Delta_\textit{MLP}\)
is, the more important text attributes are. As presented in Table~\ref{tab:main_results}, \(\Delta_\textit{MLP}\) on \texttt{OGBN-Arxiv} is solely \(2.98\),
indicating the text attributes are more important, in comparison with the ones in \texttt{OGBN-Products} and \texttt{OGBL-Citation2}. This empirically
indicates why the performance of \ourmethod\ in \texttt{OGBN-Products} does not perform as well as the one in \texttt{OGBN-Arxiv}. We show a sample of
text in \texttt{OGBN-Arxiv} and \texttt{OGBN-Products} respectively in Appendix~\ref{sec:dataset_details}. We find that the text in \texttt{OGBN-products}
resembles more a bag of words, which account for the less improvement when using LM features.

\subsection{\textbf{Q2}: Does \textbf{X}-\ourmethod\ facilitate better convergence for GNNs?}

\begin{figure}[!t]
   \vspace{-3mm}
   \includegraphics[width=\textwidth]{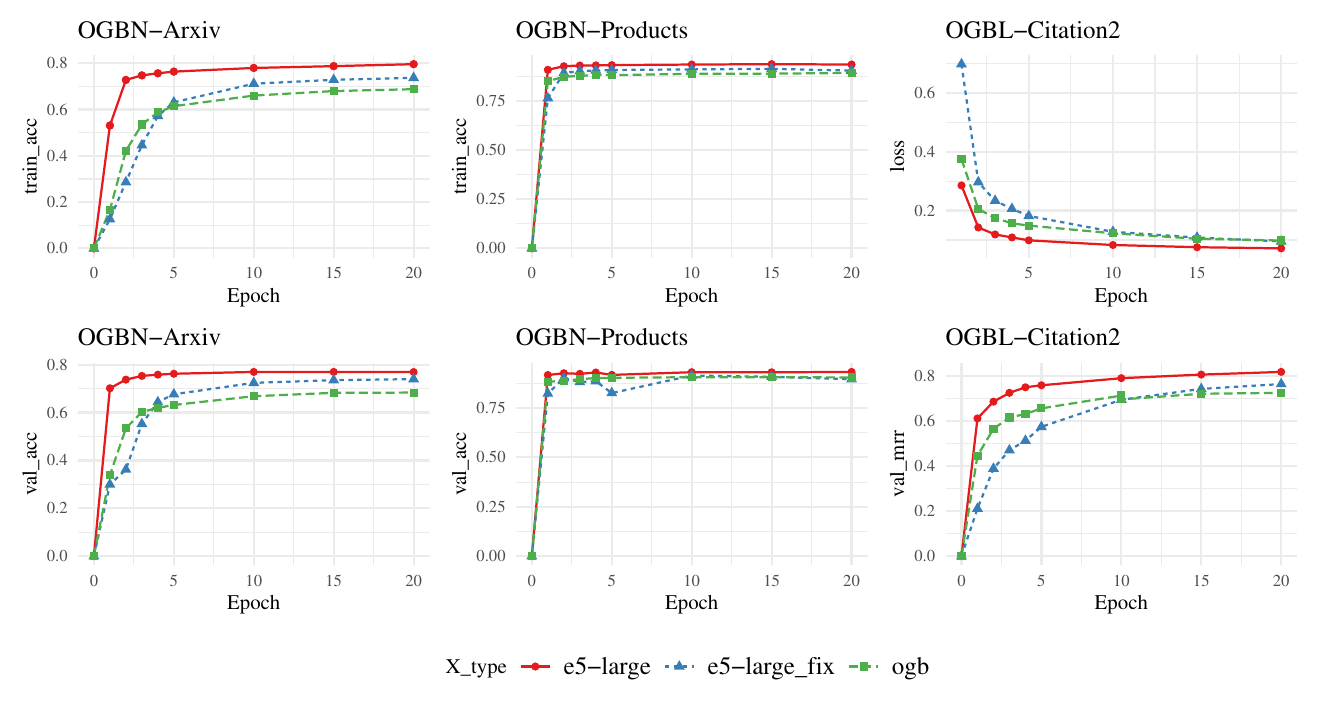}
   \vspace{-8mm}
   \caption{Training convergence and validation results of GNNs with \(\mX\)-\ourmethod,
      \(\mX\)-OGB, and \(\mX\)-FIX. The LM backbone is e5-large. The learning rate and
      batch size are consistent.}
   \label{fig:convergence}
   \vspace{-3mm}
\end{figure}

Towards a comprehensive understanding of the effectiveness of \ourmethod, we further investigate the convergence of GNNs with \ourmethod.
We compare the training convergence and the corresponding validation performance of GNNs trained on \ourmethod, \(\mX\)-OGB, and  \(\mX\)-FIX,
where \(\mX\)-FIX denotes the node embeddings generated by the pretrained LMs without finetuning. The illustration is placed in
Fig.~\ref{fig:convergence}. It is worth noting that we use the training accuracy on \texttt{OGBN-Arxiv} and \texttt{OGBN-Products} to denote their
convergence since we utilize \textit{label smoothing} during training which make the training loss not directly comparable on them. Based on
Fig.~\ref{fig:convergence}, we have the following observation:

\noindent
\observation
\textbf{\ourmethod\ moderately speeds up and stabilizes the training of GNNs.}
As shown in Fig.~\ref{fig:convergence}, GNNs with \ourmethod\ generally converge faster than the ones with \(\mX\)-OGB and \(\mX\)-FIX. With \ourmethod,
GraphSAGE could converge within 2 epochs on \texttt{OGBN-Arxiv} and \texttt{OGBN-Products}. In contrast, training on the features directly generated by
the pretrained LMs (i.e., \(\mX\)-FIX) converges much slower, even slower than one of \(\mX\)-OGB (possibly due to a larger hidden dimension). This further
indicates the benefits of \ourmethod.

\vspace{-2mm}
\subsection{\textbf{Q3}: Is PEFT a necessity for LM finetuning stage?}\label{sec:ablation_peft}

In this ablation study, we analyze the effectiveness of PEFT for LM finetuning stage in \ourmethod. Besides the accelerating finetuning,
we also find notable contribution of PEFT to the effectiveness. We summarize the training,
validation, and test accuracy of two stages: LM finetuning stage and GNN training stage. The results of node classification
are presented in Table~\ref{tab:ablatio_peft}.

\begin{table}[!t]
   \vspace{-5mm}
   \centering
   \caption{The training results of finetuning LM (\textit{LM stage}) and further training GNN on top of derived features (\textit{GNN stage}).
      We compare the results of PEFT (\ourmethod) with full-finetuning (\(\mX\)-FULL). The LM backbone is e5-large and the GNN backbone is GraphSAGE.
      We \textbf{bold} the better results on each comparison.
      \(\Delta_\textit{overfit}\) computes (Train Acc. - Test Acc.) to measure the overfitting.}\label{tab:ablatio_peft}
   \resizebox{0.8\textwidth}{!}{
      \begin{tabular}{lll|ccc|g}
         \toprule
         datasets                  & Stage                & X\_type            & Train Acc. & Valid Acc & Test Acc.      & \(\Delta_\textit{overfit}\) \\
         \midrule
         \multirow{4}{*}{Arxiv}    & \multirow{2}{*}{LM}  & \(\mX\)-FULL       & 82.33      & 75.85     & \textbf{74.77} & 7.56                        \\
                                   &                      & \(\mX\)-\ourmethod & 75.72      & 75.40     & 74.31          & 1.41                        \\
         \cmidrule(lr){2-7}
                                   & \multirow{2}{*}{GNN} & \(\mX\)-FULL       & 84.39      & 76.73     & 75.28          & 9.11                        \\
                                   &                      & \(\mX\)-\ourmethod & 79.37      & 77.47     & \textbf{76.85} & 2.52                        \\
         \midrule
         \multirow{4}{*}{Products} & \multirow{2}{*}{LM}  & \(\mX\)-FULL       & 95.46      & 91.70     & \textbf{78.70} & 16.76                       \\
                                   &                      & \(\mX\)-\ourmethod & 89.45      & 88.85     & 77.81          & 11.64                       \\
         \cmidrule(lr){2-7}
                                   & \multirow{2}{*}{GNN} & \(\mX\)-FULL       & 96.42      & 93.18     & 81.80          & 14.62                       \\
                                   &                      & \(\mX\)-\ourmethod & 95.37      & 93.57     & \textbf{84.58} & 10.79                       \\
         \bottomrule
      \end{tabular}
   }
\end{table}
\noindent
\observation
\textbf{PEFT could significantly alleviate the overfitting problem during finetuning LM and further facilitate
   the training of GNNs with regularized features.}
As shown in Table~\ref{tab:ablatio_peft}, due to the excessively strong learning capacity of LMs, finetuning LMs on the downstream
task causes a severe overfitting problem. Although full-finetuning outperforms PEFT in LM stage, training GNNs on the
derived features gains notably less improvement. In contrast, PEFT could significantly mitigate the overfitting issue according to
\(\Delta_\textit{overfit}\) in LM finetuning stage and assist the training of GNNs with regularized features to gain considerable
improvement compared with full-finetuning.

\vspace{-2mm}
\subsection{\textbf{Q4}: How sensitive is GNN training sensitive to the selection of LMs?}\label{sec:ablation_lm}

In this experiment, we investigate the effects of the selection of LMs. In detail, we aim at answering two questions:
\((i)\) Is the training of GNN sensitive to the selection of LMs? \((ii)\) is retrieval LM a better choice than generally
pretrained LMs, e.g. masked language modeling (MLM)?
To answer the above questions, we conduct experiments on multiple LM backbones. Particularly, to answer the second question,
besides the three retrieval LMs introduced in Table~\ref{tab:lm_stats},
we also consider a LM pretrained with MLM, i.e., \textit{roberta-large}~\citep{liu2019roberta}, which
has exactly the same architecture with \textit{all-roberta-large-v1}. Based on the results, we have the following observations.

\begin{table}[!t]
   \vspace{-5mm}
   \centering
   \caption{The performance of GraphSAGE and MLP trained on \ourmethod with different LM backbones.
      The former three LMs are finetuned with searched hyperparameters. For each row,
      we \textbf{bold} the best result and \underline{underline} the runner-up. all results are reported
      based on 10 runs.}\label{tab:ablation_lm}
   \resizebox{0.9\textwidth}{!}{
      \begin{tabular}{lll|ccc}
         \toprule
         \multirow{2}{*}{datasets}  & \multirow{2}{*}{Metric} & LM backbone & all-MiniLM-L6-v2   & all-roberta-large-v1           & e5-large                       \\
         \cmidrule(lr){3-6}
                                    &                         & \#Params.   & 22M                & 355M                           & 335M                           \\
         \midrule
         \multirow{2}{*}{Arxiv}     & \multirow{2}{*}{Acc.}   & MLP         & 70.56 \(\pm\) 0.09 & 74.32 \(\pm\) 0.12             & 74.06 \(\pm\) 0.13             \\
                                    &                         & GraphSAGE   & 75.14 \(\pm\) 0.30 & \underline{76.18 \(\pm\) 0.37} & \textbf{76.84 \(\pm\) 0.34}    \\
         \midrule
         \multirow{2}{*}{Products}  & \multirow{2}{*}{Acc.}   & MLP         & 72.36 \(\pm\) 0.12 & 77.48 \(\pm\) 0.19             & 76.73 \(\pm\) 0.44             \\
                                    &                         & GraphSAGE   & 82.04 \(\pm\) 0.57 & \underline{83.68 \(\pm\) 0.32} & \textbf{84.59 \(\pm\) 0.44}    \\
         \midrule
         \multirow{2}{*}{Citation2} & \multirow{2}{*}{MRR}    & MLP         & 64.49 \(\pm\) 0.18 & 70.32 \(\pm\) 0.22             & 72.90 \(\pm\) 0.14             \\
                                    &                         & GraphSAGE   & 83.09 \(\pm\) 0.75 & \textbf{85.29 \(\pm\) 0.70}    & \underline{85.13 \(\pm\) 0.73} \\
         \bottomrule
      \end{tabular}
   }
   \vspace{-3mm}
\end{table}

\noindent
\observation
\textbf{GNN's training is moderately sensitive to the selection of LMs.}
We select three retrieval LMs based on their rank in MTEB leaderboard in terms of the classification and retrieval performance.
Interestingly, based on the leaderboard, the performance ranking is \textit{e5-large} \(>\) \textit{all-roberta-large-v1}
\(>\) \textit{all-MiniLM-L6-v2}, which is consistent with their overall performance in Table~\ref{tab:ablation_lm}. We conjecture
that with more powerful retrieval LMs, the performance of GNNs will be further improved. In addition, we perform an ablation study
regarding the comparison between retrieval LMs and MLM LMs. The results are shown in Table~\ref{tab:ablation_retrieval}. We observe
that given the same architecture, the models specifically pretrained for retrieval tasks (\textit{all-roberta-large-v1}) generally perform better on tasks of TG representation learning.

\vspace{-2mm}
\section{Conclusion}~\label{sec:conclusion}
\noindent
In this work, we propose a frustratingly simple, yet highly effective approach \ourmethod\ for TG representation learning. We show that with
a parameter-efficiently finetuned LM on the same downstream task first, a simple two-layer
GraphSAGE trained on the generated node embeddings can achieve on-par state-of-the-art (SOTA) performance on
\texttt{OGBN-Arxiv} (77.48 \%). Furthermore, with SOTA GNN, we achieve new SOTA of \(78.04\%\) on \texttt{OGBN-Arxiv}.
It is worth noting this simple baseline complements any LMs and GNNs, and we expect further performance improvement given more powerful LMs and GNNs.

\bibliography{iclr2023_conference}
\bibliographystyle{iclr2023_conference}

\clearpage
\appendix
\renewcommand{\thepage}{A\arabic{page}}
\renewcommand{\thesection}{A\arabic{section}}
\renewcommand{\thetable}{A\arabic{table}}
\renewcommand{\thefigure}{A\arabic{figure}}

\section{More Experiment Results}\label{sec:more_exp_results}

\begin{table}[!ht]
    \centering
    \caption{Node Classification Accuracy of \(\mathbf{X}\)-\ourmethod\ on ogbn-arxiv (Arxiv) and ogbn-products (Products).
        All reported results are averaged over \(10\) runs in the format of mean \(\pm\) std.
        We \colorbox{red!25}{red color} the best results and \colorbox{blue!25}{blue color} the runner-ups with the same GNN backbone.
        \(\uparrow (\%)\) denotes the improvement of \(\mathbf{X}\)-\ourmethod\ over the original feature \(\mX\)-OGB.
        \(\Delta_\textit{MLP}\) and \(\Delta_\textit{GNN}\) denotes the extreme value difference among all methods (including MLP) and GNNs, respectively.}
    \label{tab:node_cls_results_main}
    \resizebox{1\linewidth}{!}{
        \begin{tabular}{ll|c|ccc|cc|cc|cc}
            \toprule
            \multirow{2}{*}{Datasets} & \multirow{2}{*}{GNN}                                                & \multirow{2}{*}{Acc. (\%)}                             & \multicolumn{3}{c}{Baselines}                         & \multicolumn{6}{c}{\(\mathbf{X}\)-\ourmethod}                                                                                                                                                                                                                                                                                                   \\
            \cmidrule(lr){4-6} \cmidrule(lr){7-12}
                                      &                                                                     &                                                        & \(\mathbf{X}\)-OGB                                    & \(\mathbf{X}\)-GIANT                          & GLEM\(^a\)                                             & MiniLM-L6          & \(\uparrow\) (\%)                                     & e5-large                              & \(\uparrow\) (\%)                                     & roberta-large                         & \(\uparrow\) (\%) \\

            \midrule
            \multirow{9}{*}{Arxiv}    & \multirow{2}{*}{MLP}                                                & \textit{val}                                           & 49.14 \(\pm\) 0.27                                    & 72.02 \(\pm\) 0.16                            & -                                                      & 71.59 \(\pm\) 0.07 & 22.45                                                 & 75.08 \(\pm\) 0.09                    & 26.66                                                 & 74.80 \(\pm\) 0.07                    & 25.66             \\
                                      &                                                                     & \textit{test}                                          & 47.73 \(\pm\) 0.29                                    & 71.08 \(\pm\) 0.22                            & -                                                      & 70.56 \(\pm\) 0.09 & 22.83                                                 & \cellcolor{blue!25}74.06 \(\pm\) 0.13 & 26.33                                                 & \cellcolor{red!25}74.32 \(\pm\) 0.12  & 26.59             \\
                                      & \multirow{2}{*}{GraphSAGE}                                          & \textit{val}                                           & 72.80 \(\pm\) 0.18                                    & 74.58 \(\pm\) 0.20                            & 76.45 \(\pm\) 0.05                                     & 75.92 \(\pm\) 0.17 & 3.12                                                  & 77.47 \(\pm\) 0.14                    & 4.67                                                  & 76.86 \(\pm\) 0.13                    & 4.06              \\
                                      &                                                                     & \textit{test}                                          & 71.80 \(\pm\) 0.20                                    & 73.70 \(\pm\) 0.09                            & 75.50 \(\pm\) 0.24                                     & 75.14 \(\pm\) 0.30 & 3.34                                                  & \cellcolor{red!25}76.84 \(\pm\) 0.34  & 5.04                                                  & \cellcolor{blue!25}76.18 \(\pm\) 0.37 & 4.38              \\
                                      & \multirow{2}{*}{GAMLP}                                              & \textit{val}                                           & 71.49 \(\pm\) 0.41                                    & 76.36 \(\pm\) 0.09                            & 76.95 \(\pm\) 0.14                                     & 76.75 \(\pm\) 0.11 & 5.26                                                  & 77.90 \(\pm\) 0.12                    & 6.41                                                  & 77.57 \(\pm\) 0.15                    & 6.08              \\
                                      &                                                                     & \textit{test}                                          & 70.61 \(\pm\) 0.52                                    & 75.26 \(\pm\) 0.15                            & 75.62 \(\pm\) 0.23                                     & 75.46 \(\pm\) 0.17 & 4.85                                                  & \cellcolor{red!25}76.92 \(\pm\) 0.10  & 6.31                                                  & \cellcolor{blue!25}76.72 \(\pm\) 0.19 & 6.11              \\
                                      & \multirow{2}{*}{SAGN}                                               & \textit{val}                                           & 72.74 \(\pm\) 0.39                                    & 75.76 \(\pm\) 0.21                            & -                                                      & 76.84 \(\pm\) 0.08 & 4.10                                                  & 78.03 \(\pm\) 0.05                    & 5.29                                                  & 77.63 \(\pm\) 0.16                    & 4.89              \\
                                      &                                                                     & \textit{test}                                          & 71.76 \(\pm\) 0.41                                    & 74.39 \(\pm\) 0.38                            & -                                                      & 75.50 \(\pm\) 0.23 & 3.74                                                  & \cellcolor{red!25}76.85 \(\pm\) 0.12  & 5.09                                                  & \cellcolor{blue!25}76.59 \(\pm\) 0.17 & 4.83              \\
                                      & \multirow{2}{*}{RevGAT}                                             & \textit{val}                                           & 75.10 \(\pm\) 0.15                                    & 76.97 \(\pm\) 0.08                            & 77.49 \(\pm\) 0.17                                     & 76.86 \(\pm\) 0.24 & 1.76                                                  & 77.68 \(\pm\) 0.07                    & 2.58                                                  & 76.32 \(\pm\) 0.18                    & 1.22              \\
                                      &                                                                     & \textit{test}                                          & 74.01 \(\pm\) 0.29                                    & 75.93 \(\pm\) 0.22                            & \cellcolor{blue!25}76.97 \(\pm\) 0.19                  & 75.96 \(\pm\) 0.21 & 1.95                                                  & \cellcolor{red!25}77.04 \(\pm\) 0.13  & 3.03                                                  & 75.88 \(\pm\) 0.58                    & 1.87              \\
            \cmidrule(lr){2-12}
                                      & \multicolumn{2}{c}{\(\Delta_\textit{MLP}\)/\(\Delta_\textit{GNN}\)} & \textcolor{pigment}{25.24} / \textcolor{pigment}{3.40} & \textcolor{pigment}{4.85} / \textcolor{pigment}{2.23} & -                                             & \textcolor{pigment}{5.40} / \textcolor{pigment}{0.82}  & -                  & 2.98 / \textcolor{pigment}{0.20}                      & -                                     & 2.40 / \textcolor{pigment}{0.84}                      & -                                                         \\
            \midrule
            \multirow{8}{*}{Products} & \multirow{2}{*}{MLP}                                                & \textit{val}                                           & 63.44 \(\pm\) 0.30                                    & 89.67 \(\pm\) 0.07                            & -                                                      & 86.82 \(\pm\) 0.02 & 23.38                                                 & 88.75 \(\pm\) 0.04                    & 25.31                                                 & 90.01 \(\pm\) 0.03                    & 26.57             \\
                                      &                                                                     & \textit{test}                                          & 50.86 \(\pm\) 0.26                                    & \cellcolor{red!25}77.58 \(\pm\) 0.24          & -                                                      & 72.36 \(\pm\) 0.12 & 21.50                                                 & 76.73 \(\pm\) 0.44                    & 25.87                                                 & \cellcolor{blue!25}77.48 \(\pm\) 0.19 & 26.62             \\
                                      & \multirow{2}{*}{GraphSAGE}                                          & \textit{val}                                           & 90.03 \(\pm\) 0.08                                    & 93.49 \(\pm\) 0.09                            & 93.84 \(\pm\) 0.12                                     & 93.49 \(\pm\) 0.08 & 3.46                                                  & 93.57 \(\pm\) 0.20                    & 3.54                                                  & 93.34 \(\pm\) 0.09                    & 3.31              \\
                                      &                                                                     & \textit{test}                                          & 78.81 \(\pm\) 0.23                                    & 82.84 \(\pm\) 0.29                            & 83.16 \(\pm\) 0.19                                     & 82.04 \(\pm\) 0.57 & 3.23                                                  & \cellcolor{red!25}84.59 \(\pm\) 0.44  & 5.78                                                  & \cellcolor{blue!25}83.68 \(\pm\) 0.32 & 4.87              \\
                                      & \multirow{2}{*}{SAGN+SCR}                                           & \textit{val}                                           & 91.83 \(\pm\) 0.24                                    & 94.04 \(\pm\) 0.12                            & 94.00 \(\pm\) 0.03                                     & 92.89 \(\pm\) 0.07 & 1.06                                                  & 94.12 \(\pm\) 0.10                    & 2.29                                                  & 94.13 \(\pm\) 0.12                    & 2.30              \\
                                      &                                                                     & \textit{test}                                          & 81.82 \(\pm\) 0.44                                    & \cellcolor{blue!25}86.12 \(\pm\) 0.34         & \cellcolor{red!25}87.36 \(\pm\) 0.07                   & 82.43 \(\pm\) 0.40 & 0.61                                                  & 85.40 \(\pm\) 0.28                    & 3.58                                                  & 85.23 \(\pm\) 0.32                    & 3.41              \\
            \cmidrule(lr){2-12}
                                      & \multicolumn{2}{c}{\(\Delta_\textit{MLP}\)/\(\Delta_\textit{GNN}\)} & \textcolor{pigment}{30.96} / \textcolor{pigment}{3.01} & \textcolor{pigment}{8.54} / \textcolor{pigment}{3.28} & -                                             & \textcolor{pigment}{10.07} / \textcolor{pigment}{0.39} &                    & \textcolor{pigment}{8.67} / \textcolor{pigment}{0.81} & -                                     & \textcolor{pigment}{7.75} / \textcolor{pigment}{1.55} & -                                                         \\
            \bottomrule
            \multicolumn{10}{l}{\(^a\) results are from the original papers.}                                                                                                                                                                                                                                                                                                                                                                                                                                                                                                  \\
            \multicolumn{10}{l}{}                                                                                                                                                                                                                                                                                                                                                                                                                                                                                                                                              \\
        \end{tabular}}
\end{table}

\begin{table}[!ht]
    \centering
    \caption{Link prediction results on \textit{OGBL-Citation2-2.7M} (Citation2). All reported results are
        averaged over 5 runs. We \colorbox{red!25}{red color} the best results and \colorbox{blue!25}{blue color} the runner-ups with the same GNN backbone.
        \(\uparrow (\%)\) denotes the improvement of \(\mathbf{X}\)-\ourmethod\ over the original feature \(\mX\)-OGB.
        \(\Delta_\textit{MLP}\) and \(\Delta_\textit{GNN}\) denotes the margin of (MLP, SEAL) and (GraphSAGE, SEAL), respectively. We use \textcolor{pigment}{blue}
        color denoting the negative values and \textcolor{Brown}{red} denoting positive. Specifically, in the context of \(\Delta\), positives indicate MLP/GraphSAGE
        performs better than SEAL.}\label{tab:link_pred_results_main}
    \resizebox{1\textwidth}{!}{
        \begin{tabular}{lll|c|cc|cc|cc}
            \toprule
            \multirow{2}{*}{Metrics}          & \multirow{2}{*}{GNN}                                                & \multirow{2}{*}{Split}                                    & Baselines                                                & \multicolumn{6}{c}{\(\mX\)-\ourmethod}                                                                                                                                                                                                                           \\
            \cmidrule(lr){4-4} \cmidrule(lr){5-10}
                                              &                                                                     &                                                           & \(\mX\)-OGB                                              & MiniLM-L6                              & \(\uparrow (\%)\)                                        & roberta-large                         & \(\uparrow (\%)\)                                        & e5-large                              & \(\uparrow (\%)\) \\
            \midrule
            \multirow{6}{*}{\textit{MRR}}     & \multirow{2}{*}{MLP}                                                & \textit{val}                                              & 25.37 \(\pm\) 0.09                                       & 64.56 \(\pm\) 0.15                     & 39.19                                                    & 70.20 \(\pm\) 0.19                    & 44.83                                                    & 72.79 \(\pm\) 0.17                    & 47.42             \\
                                              &                                                                     & \textit{test}                                             & 25.44 \(\pm\) 0.01                                       & 64.49 \(\pm\) 0.18                     & 39.05                                                    & \cellcolor{blue!25}70.32 \(\pm\) 0.22 & 44.88                                                    & \cellcolor{red!25}72.90 \(\pm\) 0.14  & 47.46             \\
                                              & \multirow{2}{*}{GraphSAGE}                                          & \textit{val}                                              & 77.40 \(\pm\) 0.88                                       & 83.13 \(\pm\) 0.72                     & 5.73                                                     & 85.27 \(\pm\) 0.78                    & 7.87                                                     & 85.20 \(\pm\) 0.69                    & 7.80              \\
                                              &                                                                     & \textit{test}                                             & 77.31 \(\pm\) 0.90                                       & 83.09 \(\pm\) 0.75                     & 5.78                                                     & \cellcolor{red!25}85.29 \(\pm\) 0.70  & 7.98                                                     & \cellcolor{blue!25}85.13 \(\pm\) 0.73 & 7.82              \\
                                              & \multirow{2}{*}{SEAL}                                               & \textit{val}                                              & 87.21 \(\pm\) 0.03                                       & 88.33 \(\pm\) 0.30                     & 1.12                                                     & 88.29 \(\pm\) 0.45                    & 1.08                                                     & 88.56 \(\pm\) 0.38                    & 1.35              \\
                                              &                                                                     & \textit{test}                                             & 86.14 \(\pm\) 0.40                                       & \cellcolor{blue!25}86.69 \(\pm\) 0.43  & 0.55                                                     & \cellcolor{red!25}87.02 \(\pm\) 0.46  & 0.88                                                     & 86.66 \(\pm\) 1.21                    & 0.52              \\
            \cmidrule(lr){2-10}
                                              & \multicolumn{2}{c}{\(\Delta_\textit{MLP}\)/\(\Delta_\textit{GNN}\)} & \textcolor{pigment}{-60.70} / \textcolor{pigment}{-8.83}  & \textcolor{pigment}{-22.20} / \textcolor{pigment}{-3.60} & -                                      & \textcolor{pigment}{-16.70} /\textcolor{pigment}{-1.73}  & -                                     & \textcolor{pigment}{-13.76} / \textcolor{pigment}{-1.53} & -                                                         \\
            \midrule
            \multirow{6}{*}{\textit{Hits@1}}  & \multirow{2}{*}{MLP}                                                & \textit{val}                                              & 15.04 \(\pm\) 0.09                                       & 52.29 \(\pm\) 0.18                     & 37.25                                                    & 59.46 \(\pm\) 0.19                    & 44.42                                                    & 62.21 \(\pm\) 0.23                    & 47.17             \\
                                              &                                                                     & \textit{test}                                             & 15.11 \(\pm\) 0.06                                       & 52.18 \(\pm\) 0.25                     & 37.07                                                    & \cellcolor{blue!25}59.66 \(\pm\) 0.26 & 44.55                                                    & \cellcolor{red!25}62.31 \(\pm\) 0.19  & 47.20             \\
                                              & \multirow{2}{*}{GraphSAGE}                                          & \textit{val}                                              & 67.28 \(\pm\) 1.20                                       & 74.83 \(\pm\) 1.02                     & 7.55                                                     & 77.98 \(\pm\) 1.20                    & 10.70                                                    & 77.73 \(\pm\) 0.89                    & 10.45             \\
                                              &                                                                     & \textit{test}                                             & 67.09 \(\pm\) 1.25                                       & 74.79 \(\pm\) 1.10                     & 7.70                                                     & \cellcolor{red!25}77.99 \(\pm\) 0.89  & 10.90                                                    & \cellcolor{blue!25}77.66 \(\pm\) 0.91 & 10.57             \\
                                              & \multirow{2}{*}{SEAL}                                               & \textit{val}                                              & 82.76 \(\pm\) 0.14                                       & 84.35 \(\pm\) 0.42                     & 1.59                                                     & 84.25 \(\pm\) 0.79                    & 1.49                                                     & 84.70 \(\pm\) 0.58                    & 1.94              \\
                                              &                                                                     & \textit{test}                                             & \cellcolor{blue!25}81.74 \(\pm\) 0.46                    & 81.40 \(\pm\) 0.96                     & -0.34                                                    & \cellcolor{red!25}82.34 \(\pm\) 0.79  & 0.60                                                     & 81.15 \(\pm\) 2.04                    & -0.59             \\
            \cmidrule(lr){2-10}
                                              & \multicolumn{2}{c}{\(\Delta_\textit{MLP}\)/\(\Delta_\textit{GNN}\)} & \textcolor{pigment}{-66.63} / \textcolor{pigment}{-14.65} & \textcolor{pigment}{-29.22} /\textcolor{pigment}{-6.61}  & -                                      & \textcolor{pigment}{-22.68} / \textcolor{pigment}{-4.35} & -                                     & \textcolor{pigment}{-18.84} / \textcolor{pigment}{-3.39} & -                                                         \\
            \midrule
            \multirow{6}{*}{\textit{Hits@3}}  & \multirow{2}{*}{MLP}                                                & \textit{val}                                              & 28.06 \(\pm\) 0.10                                       & 72.60 \(\pm\) 0.16                     & 44.54                                                    & 77.56 \(\pm\) 0.23                    & 49.50                                                    & 80.42 \(\pm\) 0.15                    & 52.36             \\
                                              &                                                                     & \textit{test}                                             & 28.22 \(\pm\) 0.02                                       & 72.62 \(\pm\) 0.19                     & 44.40                                                    & \cellcolor{blue!25}77.66 \(\pm\) 0.24 & 49.44                                                    & \cellcolor{red!25}80.55 \(\pm\) 0.13  & 52.33             \\
                                              & \multirow{2}{*}{GraphSAGE}                                          & \textit{val}                                              & 85.54 \(\pm\) 0.69                                       & 90.17 \(\pm\) 0.61                     & 4.63                                                     & 91.55 \(\pm\) 0.98                    & 6.01                                                     & 91.72 \(\pm\) 0.90                    & 6.18              \\
                                              &                                                                     & \textit{test}                                             & 85.56 \(\pm\) 0.69                                       & 90.16 \(\pm\) 0.51                     & 4.60                                                     & \cellcolor{blue!25}91.57 \(\pm\) 1.10 & 6.01                                                     & \cellcolor{red!25}91.62 \(\pm\) 0.87  & 6.06              \\
                                              & \multirow{2}{*}{SEAL}                                               & \textit{val}                                              & 91.36 \(\pm\) 0.44                                       & 92.00 \(\pm\) 0.07                     & 0.64                                                     & 92.15 \(\pm\) 0.19                    & 0.79                                                     & 91.75 \(\pm\) 0.18                    & 0.39              \\
                                              &                                                                     & \textit{test}                                             & 90.92 \(\pm\) 0.32                                       & 91.42 \(\pm\) 0.60                     & 0.50                                                     & \cellcolor{red!25}91.52 \(\pm\) 0.56  & 0.60                                                     & \cellcolor{blue!25}91.42 \(\pm\) 0.19 & 0.50              \\
            \cmidrule(lr){2-10}
                                              & \multicolumn{2}{c}{\(\Delta_\textit{MLP}\)/\(\Delta_\textit{GNN}\)} & \textcolor{pigment}{-62.70} / \textcolor{pigment}{-5.36}  & \textcolor{pigment}{-18.80} / \textcolor{pigment}{-1.26} & -                                      & \textcolor{pigment}{-13.86} / \textcolor{Brown}{0.05}    & -                                     & \textcolor{pigment}{-10.87} / \textcolor{Brown}{0.20}    & -                                                         \\
            \midrule
            \multirow{6}{*}{\textit{Hits@10}} & \multirow{2}{*}{MLP}                                                & \textit{val}                                              & 46.73 \(\pm\) 0.14                                       & 87.62 \(\pm\) 0.06                     & 40.89                                                    & 89.80 \(\pm\) 0.20                    & 43.07                                                    & 91.74 \(\pm\) 0.08                    & 45.01             \\
                                              &                                                                     & \textit{test}                                             & 46.59 \(\pm\) 0.11                                       & 87.57 \(\pm\) 0.12                     & 40.98                                                    & \cellcolor{blue!25}89.66 \(\pm\) 0.14 & 43.07                                                    & \cellcolor{red!25}91.74 \(\pm\) 0.10  & 45.15             \\
                                              & \multirow{2}{*}{GraphSAGE}                                          & \textit{val}                                              & 94.29 \(\pm\) 0.19                                       & 96.25 \(\pm\) 0.13                     & 1.96                                                     & 96.61 \(\pm\) 0.12                    & 2.32                                                     & 96.71 \(\pm\) 0.09                    & 2.42              \\
                                              &                                                                     & \textit{test}                                             & 94.37 \(\pm\) 0.17                                       & 96.30 \(\pm\) 0.13                     & 1.93                                                     & \cellcolor{blue!25}96.64 \(\pm\) 0.12 & 2.27                                                     & \cellcolor{red!25}96.74 \(\pm\) 0.11  & 2.37              \\
                                              & \multirow{2}{*}{SEAL}                                               & \textit{val}                                              & 94.59 \(\pm\) 0.14                                       & 94.88 \(\pm\) 0.25                     & 0.29                                                     & 95.08 \(\pm\) 0.12                    & 0.49                                                     & 95.08 \(\pm\) 0.21                    & 0.49              \\
                                              &                                                                     & \textit{test}                                             & 93.90 \(\pm\) 0.49                                       & \cellcolor{blue!25}94.40 \(\pm\) 0.07  & 0.50                                                     & 93.95 \(\pm\) 0.37                    & 0.05                                                     & \cellcolor{red!25}94.54 \(\pm\) 0.25  & 0.64              \\
            \cmidrule(lr){2-10}
                                              & \multicolumn{2}{c}{\(\Delta_\textit{MLP}\)/\(\Delta_\textit{GNN}\)} & \textcolor{pigment}{-47.31} / \textcolor{Brown}{-0.47}    & \textcolor{pigment}{-6.83} /\textcolor{Brown}{1.90}      & -                                      & \textcolor{pigment}{-4.29} / \textcolor{Brown}{2.66}     & -                                     & \textcolor{pigment}{-2.80} / \textcolor{Brown}{2.20}     & -                                                         \\
            \bottomrule
        \end{tabular}
    }

\end{table}

\begin{table}[!ht]
    \centering
    \vspace{-5mm}
    \caption{The performance of Graph and MLP trained on \ourmethod\ backed with
        all-roberta-large-v1 and roberta-large, which have the same model architecture
        and differs from the pretraining strategy. we \textbf{bold} the best results
        for each comparison in \(\mX\)-Fix and \(\mX\)-\ourmethod.
        all results are reported based on 10 runs}\label{tab:ablation_retrieval}
    \resizebox{1\textwidth}{!}{
        \begin{tabular}{lll|cc|cc}
            \toprule
            \multirow{2}{*}{datasets}  & \multirow{2}{*}{Metric} & X\_type     & \multicolumn{2}{c}{\(\mX\)-Fix} & \multicolumn{2}{c}{\(\mX\)-\ourmethod}                                                    \\
            \cmidrule(lr){3-7}
                                       &                         & LM Backbone & roberta-large                   & all-roberta-large-v1                   & roberta-large      & all-roberta-large-v1        \\
            \midrule
            \multirow{2}{*}{Arxiv}     & \multirow{2}{*}{Acc.}   & MLP         & 61.15 \(\pm\) 0.83              & \textbf{72.58 \(\pm\) 0.25}            & 71.55 \(\pm\) 0.24 & \textbf{74.32 \(\pm\) 0.12} \\
                                       &                         & GraphSAGE   & 72.15 \(\pm\) 0.59              & \textbf{75.51 \(\pm\) 0.23}            & 75.48 \(\pm\) 0.16 & \textbf{76.18 \(\pm\) 0.37} \\
            \midrule
            \multirow{2}{*}{Products}  & \multirow{2}{*}{Acc.}   & MLP         & 68.14 \(\pm\) 0.23              & \textbf{70.10 \(\pm\) 0.08}            & 78.45 \(\pm\) 0.14 & \textbf{77.48 \(\pm\) 0.19} \\
                                       &                         & GraphSAGE   & 77.65 \(\pm\) 0.34              & \textbf{82.38 \(\pm\) 0.60}            & 83.56 \(\pm\) 0.21 & \textbf{83.68 \(\pm\) 0.32} \\
            \midrule
            \multirow{2}{*}{Citation2} & \multirow{2}{*}{MRR}    & MLP         & 00.20 \(\pm\) 0.01              & \textbf{70.12 \(\pm\) 0.12}            & 63.15 \(\pm\) 0.20 & \textbf{72.90 \(\pm\) 0.14} \\
                                       &                         & GraphSAGE   & 79.71 \(\pm\) 0.27              & \textbf{83.20 \(\pm\) 0.40}            & 84.37 \(\pm\) 0.34 & \textbf{85.13 \(\pm\) 0.73} \\
            \bottomrule
        \end{tabular}
    }
\end{table}

\vspace{-5mm}
\section{Reproducibility Statement}

To ensure the reproducibility of our experiments and benefit the community for further research, we provide
the source code at \ourcode\ and all node features of \ourmethod\
at \ourfeature. 

\vspace{-3mm}
\subsection{pseudo code of \ourmethod}\label{sec:pseudo_code}
\begin{algorithm}[H]
    \label{alg:lm_init}
    \caption{PyTorch-style code of \ourmethod. Left: \textit{node classification}; Right: \textit{link prediction}.}
    \begin{lstlisting}[language=Python]
    # f_lm: language model wrapped with PEFT method, f_mlp: mlp model, f_gnn: gnn model
    # inputs: A textual graph (adj_t (A), input_ids (T), att_mask (M)) and task-specific labels (Y)
    # outputs: node representations
    \end{lstlisting}
    \begin{minipage}{0.4\columnwidth}
        \begin{lstlisting}[language=Python]
    # Node Classification
    for T, M, Y in train_loader:
       X = f_lm(T, M)
       logits = f_mlp(X)
       loss = CrossEntropyLoss(logits, Y)
       loss.backward()
       lm_optimizer.step()
 
    with torch.no_grad():
       X = f_lm(T, M)
 
    f_mlp.reset_parameters()
    for A, X, Y in train_loader:
       X = f_gnn(A, X)
       logits = f_mlp(X)
       loss = CrossEntropyLoss(logits, Y)
       loss.backward()
       gnn_optimizer.step()
 
    \end{lstlisting}
    \end{minipage}
    \begin{minipage}{0.55\columnwidth}
        \begin{lstlisting}[language=Python]
    # Link Prediction
    for (T_src, T_dst), (M_src, M_dst), Y in train_loader:
       X_src, X_dst = f_lm((T_src, M_src), (T_dst, M_dst))
       logits = f_mlp(X_src, X_dst)
       loss = BCEWithLogitsLoss(logits, Y)
       loss.backward()
       lm_optimizer.step()
 
    with torch.no_grad():
       X = lm(T, M)
 
    f_mlp.reset_parameters()
    for A, (X_src, X_dst), Y in train_loader:
       X_src, X_dst = f_gnn(A, (X_src, X_dst))
       logits = f_mlp(X_src, X_dst)
       loss = BCEWithLogitsLoss(logits, Y)
       loss.backward()
       gnn_optimizer.step()
 
    \end{lstlisting}
    \end{minipage}
\end{algorithm}

\subsection{Details of TG Version for the three OGB datasets}\label{sec:dataset_details}

In this section, we present the details of the TG version of \texttt{OGBN-Arxiv}, \texttt{OGBN-Products}, and \texttt{OGBL-Citation2}.
The statistics of the three datasets are shown in Table~\ref{tab:dataset_stats} and the text resources are shown in Table~\ref{tab:dataset_text_url}.

\begin{table}[!ht]
   \vspace{-4mm}
   \centering
   \caption{Statistics of \texttt{OGBN-Arxiv}, \texttt{OGBN-Products}, and \texttt{OGBL-Citation2-2.7M}}
   \label{tab:dataset_stats}
   \resizebox{1\linewidth}{!}{
      \begin{tabular}{lrrrrr}
         \toprule
         Datasets                                 & \#Nodes       & \#Edges        & Avg. Degree & \#Task              & Metric     \\
         \midrule
         \texttt{OGBN-Arxiv} (Arxiv)              & \(169,343\)   & \(1,166,243\)  & 13.7        & node classification & Accuracy   \\
         \texttt{OGBN-Products} (Products)        & \(2,449,029\) & \(61,859,140\) & 50.5        & node classification & Accuracy   \\
         \texttt{OGBL-Citation2-2.7M} (Citation2) & \(2,728,032\) & \(27,731,705\) & 10.2        & link prediction     & MRR / Hits \\
         \bottomrule
      \end{tabular}
   }
\end{table}

\noindent
\textbf{\texttt{OGBN-Arxiv}.} \texttt{OGBN-Arxiv} is a directed academic graph, where node denotes papers and edge denotes directed citation.
The task is to predict the category of each paper as listed in \url{https://arxiv.org}. For its TG version, we use the same split as \citet{hu2020open}.
The text for each node is its title and abstract. We concatenate them for each node with the format of "\textit{title: \{title\}; abstract: \{abstract\}}"
as the corresponding node's text. For example, \textcolor{brown}{"title: multi view metric learning for multi view video summarization; abstract:
    Traditional methods on video summarization are designed to generate summaries for single-view video records; and thus they cannot fully exploit the
    redundancy in multi-view video records. In this paper, we present a multi-view metric learning framework for multi-view video summarization that
    combines the advantages of maximum margin clustering with the disagreement minimization criterion. ..."}

\noindent
\textbf{\texttt{OGBN-Products}}. \texttt{OGBN-Products} is a co-purchase graph, where node denotes a product on Amazon and
an edge denotes the co-purchase relationship between two products. The task is to predict the category of each product (node classification).
We follow the public split as \citet{hu2020open} and the text processing strategy of
GLEM~\citep{zhao2022learning}. For each node, the corresponding text is its item description. For example,
\textcolor{brown}{"My Fair Pastry (Good Eats Vol. 9)" "Disc 1: Flour Power (Scones; Shortcakes; Southern Biscuits; Salmon Turnovers; Fruit Tart;
    Funnel Cake; Sweet or Savory; Pte Choux) Disc 2: Super Sweets 4 (Banana Spitsville; Burned Peach Ice Cream; Chocolate Taffy; Acid Jellies;
    Peanut Brittle; Chocolate Fudge; Peanut Butter Fudge) ..."}

\noindent
\textbf{\texttt{OGBL-Citation2-2.7M}}. \texttt{OGBL-Citation2-2.7M} is a citation graph, where nodes denote
papers and edges denote the citations. The task is to predict the missing citation among papers (link prediction).
All papers are collected by the official from \textit{Mircrosoft Academic Graph} whereas the text resources are
not provided. Though MAG IDs for all papers are provided, we cannot find all corresponding text resources due to the
close of MAG project~\footnote{\url{https://www.microsoft.com/en-us/research/project/microsoft-academic-graph/}}.
Hence, we take an intersection of \texttt{OGBL-Citation2} and \texttt{OGBN-Papers100M} whose text resources are
provided by the official, and build a subgraph, namely \texttt{OGBL-Citation2-2.7M}. It contains
93\% nodes of \texttt{OGBL-Citation2} and offers a roughly on-par performance for baselines.

\begin{table}[!ht]
    \vspace{-4mm}
    \centering
    \caption{The URLs of text resources for \text{ogbn-arxiv}, \text{ogbn-products}, and \text{OGBL-Citation2}.}
    \label{tab:dataset_text_url}
    \resizebox{1\textwidth}{!}{
        \begin{tabular}{ll}
            \toprule
            Dataset                      & Text Resource URL                                                                            \\
            \midrule
            \texttt{OGBN-Arxiv}          & \url{https://snap.stanford.edu/ogb/data/misc/ogbn_arxiv/titleabs.tsv.gz}                     \\
            \texttt{OGBN-Products}       & \url{https://drive.google.com/u/0/uc?id=1gsabsx8KR2N9jJz16jTcA0QASXsNuKnN & export=download} \\
            \texttt{OGBL-Citation2-2.7M} & \url{https://drive.google.com/u/0/uc?id=19_hkbBUDFZTvQrM0oMbftuXhgz5LbIZY & export=download} \\
            \bottomrule
        \end{tabular}
    }
    \vspace{-4mm}
\end{table}

\subsection{Properties of Language Models}
\begin{table}[!ht]
   \vspace{-4mm}
   \centering
   \caption{Statistics of \texttt{OGBN-Arxiv}, \texttt{OGBN-Products}, and \texttt{OGBL-Citation2-2.7M}}
   \label{tab:lm_stats}
   \resizebox{1\linewidth}{!}{
      \begin{tabular}{lrrrrr}
         \toprule
         Datasets                                 & \#Nodes       & \#Edges        & Avg. Degree & \#Task              & Metric     \\
         \midrule
         \texttt{OGBN-Arxiv} (Arxiv)              & \(169,343\)   & \(1,166,243\)  & 13.7        & node classification & Accuracy   \\
         \texttt{OGBN-Products} (Products)        & \(2,449,029\) & \(61,859,140\) & 50.5        & node classification & Accuracy   \\
         \texttt{OGBL-Citation2-2.7M} (Citation2) & \(2,728,032\) & \(27,731,705\) & 10.2        & link prediction     & MRR / Hits \\
         \bottomrule
      \end{tabular}
   }
\end{table}

\vspace{-5mm}
\subsection{Hyperparameter Search Space}\label{sec:hp_search_space}
For language models, we design the hyperparameter (HP) search space as in Table~\ref{tab:hp_search_space}. Please note that for link prediction,
the label smoothing factor is omitted. For HP searching, we utilize optuna~\citep{akiba2019optuna} to search the best HPs for each dataset and each model. For LMs,
we take 10 trials. For GNNs, we take 20 trials. The final HP setting for LMs and GNNs are placed as shell scripts in our repository.
\begin{table}[!ht]
    \vspace{-4mm}
    \centering
    \caption{The search space of LMs and GNNs.}\label{tab:hp_search_space}
    \resizebox{0.8\linewidth}{!}{
        \begin{tabular}{lll|lll}
            \toprule
            \multicolumn{3}{c}{LM} & \multicolumn{3}{c}{GNN}                                                             \\
            hyperparameter         & search space            & type      & hyperparameter  & search space    & type      \\
            \midrule
            learning rate          & [1e-6, 1e-4]            & continual & learning rate   & [1e-4, 1e-2]    & continual \\
            weight decay           & [1e-7, 1e-4]            & continual & weight decay    & [1e-7, 1e-4]    & continual \\
            label smoothing        & [0.1, 0.7]              & continual & label smoothing & [0.1, 0.7]      & continual \\
            header dropout         & [0.1, 0.8]              & continual & dropout         & [0.1, 0.8]      & continual \\
            lora r                 & [1, 2, 4, 8]            & descrete  & num of layers   & [2, 3, 4, 6, 8] & descrete  \\
            lora alpha             & [4, 8, 16, 32]          & descrete                                                  \\
            lora dropout           & [0.1, 0.8]              & continual                                                 \\
            \bottomrule
        \end{tabular}
    }
\end{table}

\end{document}